\documentclass[final]{cvpr}

\usepackage{times}
\usepackage{epsfig}
\usepackage{graphicx}
\usepackage{amsmath}
\usepackage{amssymb}
\usepackage{etoolbox,siunitx}
\robustify\bfseries
\usepackage[percent]{overpic}

\usepackage{soul}

\newcommand{\changed}[1]{#1}

\usepackage[pagebackref=true,breaklinks=true,colorlinks,bookmarks=false]{hyperref}

\def\cvprPaperID{9986} 
\def\confYear{CVPR 2021}

\begin{document}

\title{Learning Delaunay Surface Elements for Mesh Reconstruction}

\author{Marie-Julie Rakotosaona\\
 LIX, Ecole Polytechnique, IP Paris\\
mrakotos@lix.polytechnique.fr \\
\and
Paul Guerrero\\
Adobe Research\\
guerrero@adobe.com\\
\and
Noam Aigerman \\
Adobe Research\\
aigerman@adobe.com \\
\and
Niloy Mitra\\
UCL, Adobe Research\\
n.mitra@ucl.ac.uk \\
\and
Maks Ovsjanikov\\
LIX, Ecole Polytechnique, IP Paris \\
maks@lix.polytechnique.fr
}
\maketitle
\pagestyle{empty}
\thispagestyle{empty}

\begin{abstract}

We present a method for reconstructing triangle meshes from point clouds. Existing learning-based methods for mesh reconstruction mostly generate triangles individually, making it hard to create manifold meshes. We leverage the properties of 2D Delaunay triangulations to construct a mesh from manifold surface elements. Our method first estimates local geodesic neighborhoods around each point. We then perform a 2D projection of these neighborhoods using a learned logarithmic map. A Delaunay triangulation in this 2D domain is guaranteed to produce a manifold patch, which we call a Delaunay surface element. We synchronize the local 2D projections of neighboring elements to maximize the manifoldness of the reconstructed mesh. Our results show that we achieve better overall manifoldness of our reconstructed meshes than current methods to reconstruct meshes with arbitrary topology. Our code, data and pretrained models can be found online: \href{https://github.com/mrakotosaon/dse-meshing}{https://github.com/mrakotosaon/dse-meshing}






\end{abstract}


\section{Introduction}

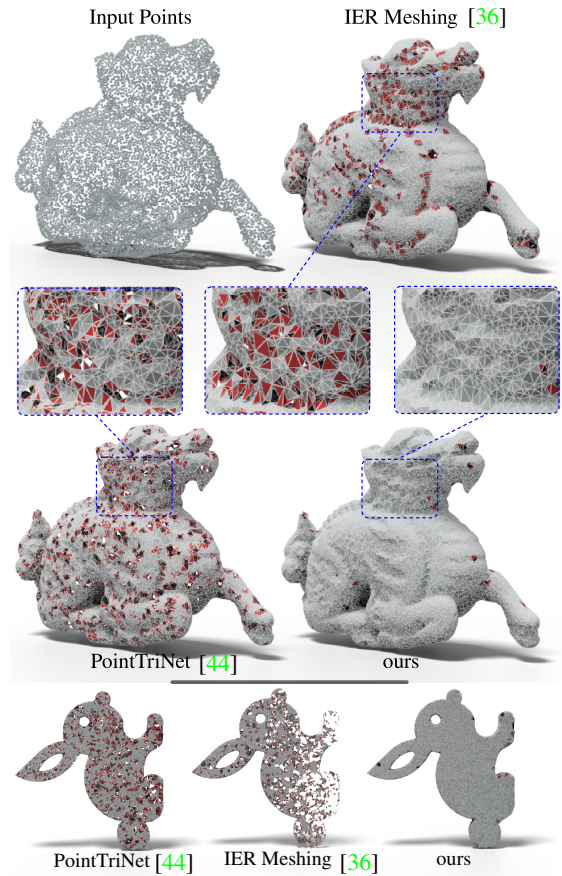
\begin{figure}[t!]
    \centering
    \begin{overpic}[width=0.965\columnwidth]{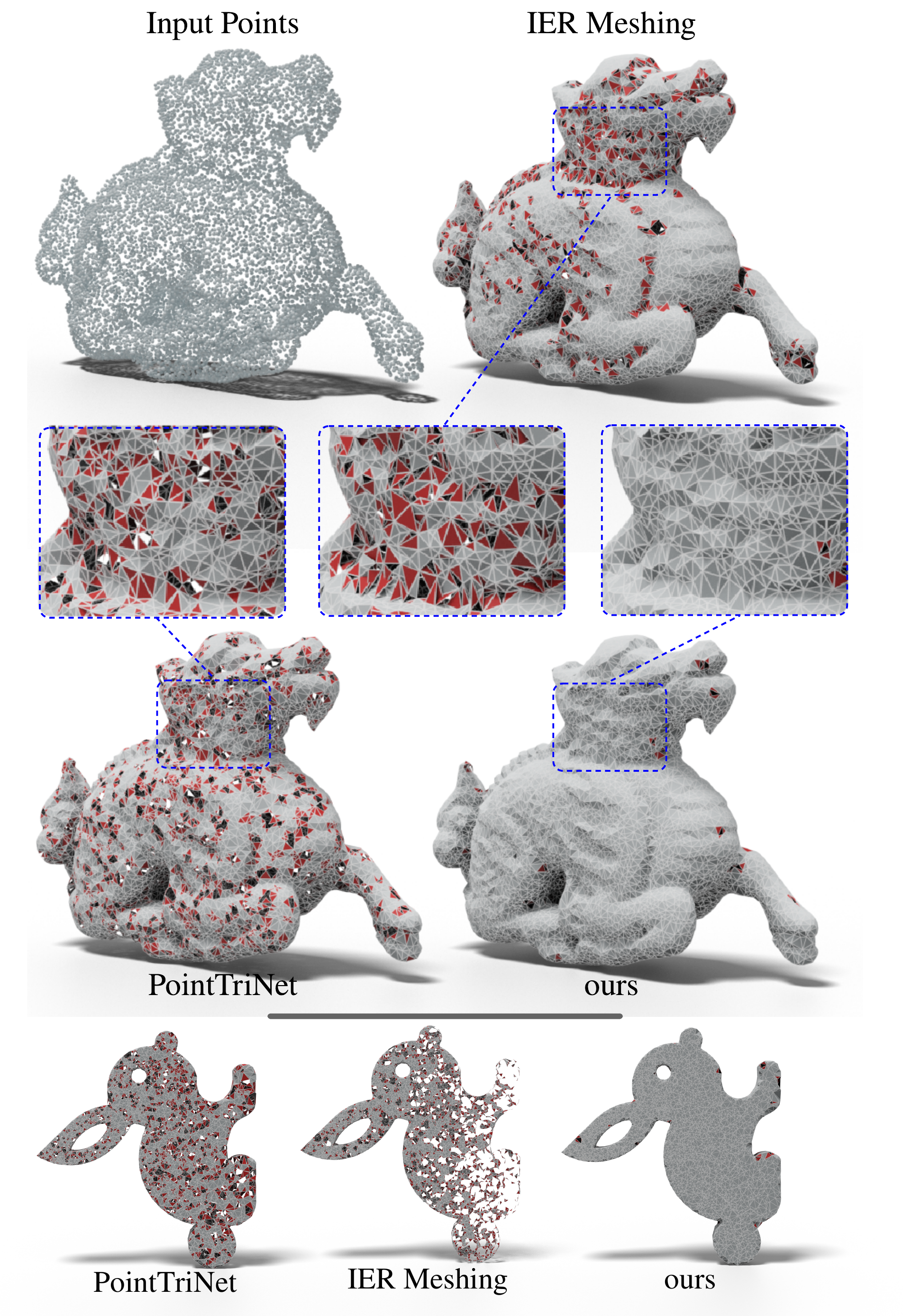}
        \put (23.5,24) {\small \cite{sharp2020pointtrinet}}
        \put (54,97.5) {\small \cite{liu2020meshing}}
        \put (18.5,1.5) {\small \cite{sharp2020pointtrinet}}
        \put (39.5,1.5) {\small \cite{liu2020meshing}}
    \end{overpic}
    \caption{We present a method for mesh reconstruction from point clouds. We combine Delaunay triangulations with learned local parameterizations to obtain a higher-quality mesh than the current state-of-the-art.
    Bad (non-manifold) triangles are shown in red. Our method is robust to uniformly (top) and non-uniformly (bottom) sampled points.  }
    \label{fig:teaser}
\end{figure}

Surface reconstruction from a given set of points (e.g., a scan), has a long history in computational geometry and computer vision \cite{berger2017survey,newman2006survey}. A version of the problem requires triangulating a given point cloud to produce a watertight and manifold surface. A key challenge is to handle different sampling conditions while producing well-shaped triangles and preserving the underlying shape features.

A good surface reconstruction algorithm should satisfy the following requirements: (i)~produce a connected, manifold and watertight triangulation; (ii)~require no case-specific parameter tuning; (iii)~preserve sharp features;  (iv)~handle point sets with non-uniform distribution; and (v)~generalize to handle a variety of shapes.

A widely-used pipeline for surface reconstruction consists in first computing an implicit surface representation \cite{kazhdan2006poisson} and then extracting a triangulation using a volumetric method such as Marching Cubes \cite{lorensen1987marching}. Methods in this category often require additional information (e.g., oriented normals), while, crucially, the resulting triangulations may not preserve the original point set and can oversmooth sharp features. On the other hand, methods from computational geometry, e.g., alpha shapes \cite{edelsbrunner1994three}, ball pivoting \cite{bernardini1999ball}, etc., can respect the original point set, come with theoretical guarantees and produce triangulations with desirable properties (e.g., good angle distribution). These approaches, however, typically require careful parameter selection and rely on dense, uniformly sampled point sets.

More recently, learning-based approaches have been developed to extract a triangulation without case-specific parameter selection.
Most of such techniques focus on robustly predicting a signed distance field or simply an occupancy map, from which a mesh is subsequently extracted using volumetric triangulation \cite{chen2019learning,genova2019learning,park2019deepsdf}.
Only two recent methods~\cite{sharp2020pointtrinet,liu2020meshing} produce a triangulation while respecting the original point set, but they ignore the quality of the triangles or have trouble reconstructing sharp features.

We present a method that combines the advantages of classical methods with learning-based data priors. Our method is based on blending together {\em Delaunay surface elements}, which are defined by a 2D Delaunay triangulation of a local neighborhood in the point set after projecting it to a planar 2D domain. For this, we propose an approach that predicts a local projection via learned logarithmic maps and uses them to propose likely triangles using local Delaunay triangulations.
%
Figure~\ref{fig:teaser} shows an example reconstructions using our method.
We evaluate our method on a benchmark of diverse surface point sets, and provide a comparison with both classical and learning-based methods to show the advantages of the proposed approach. Through these extensive experiments, we demonstrate that our method generalizes across diverse test sets, is more robust than classical approaches, and produces higher-quality triangulations than recent learning-based methods.



\section{Related Works}
\label{sec:related}


Computing a triangulation of a given point set is one of the most fundamental problems in computational geometry, computer vision, and
related disciplines. We review methods most closely related to ours and refer
to recent surveys \cite{khatamian2016survey,shewchuk2016delaunay,berger2017survey,newman2006survey} for a more in-depth discussion.

A commonly-used pipeline for surface reconstruction~\cite{hoppe1992surface,curless1996volumetric} consists of computing the implicit surface representation using, e.g., a signed distance function. A mesh can then be  extracted with standard methods such as Poisson surface reconstruction \cite{kazhdan2006poisson} combined with Marching Cubes \cite{lorensen1987marching} or Dual Contouring \cite{ju2002dual}. Such approaches work well in the presence of oriented normals and dense/uniform point sets, but do not necessarily preserve the given points in the final mesh and lead to over-smoothing or loss of details (see \cite{berger2017survey} for a detailed discussion).

\changed{We were inspired by classical methods based on Delaunay triangulations
\cite{boissonnat1984geometric,kolluri2004spectral,boissonnat2005provably,gopi2000surface,dey2001delaunay}},
alpha shapes \cite{edelsbrunner1994three} or ball pivoting \cite{bernardini1999ball}. Such approaches can be shown to recover the shape mesh topology  \cite{amenta1998new} under certain sampling conditions (an excellent overview of such approaches is provided in \cite{dey2006curve}). Unlike implicit-based methods, approaches in this category,
e.g., \cite{bernardini1999ball,amenta1999surface,boltcheva2017surface} typically preserve the input point set. However, they can often fail to produce satisfactory results for coarsely sampled shapes or in the presence of complex geometric features. Another more robust, but computationally more expensive, approach capable of feature preservation was introduced in \cite{digne2014feature}, based on iterative optimisation using optimal transport.


\subsection{Learning for surface reconstruction}
To address the challenges mentioned above, recent methods have aimed to learn surface reconstruction priors from data. The majority of existing learning-based methods in this area use a volumetric shape representation. For example, meshes can be computed by predicting voxel grid occupancy \cite{gkioxari2019mesh,mescheder2019occupancy} or via a differentiable
variant of the marching cubes~\cite{liao2018deep}, or more recently using generative models for explicit or implicit surface prediction
\cite{chen2019learning,genova2019learning,park2019deepsdf,nash2020polygen}. While these methods can produce accurate results they solve a different problem to ours and do not compute a mesh over the given point set. Instead, we focus on directly meshing a set of input points, which provides better control over the final shape and avoid over-smoothing, often associated with implicit surface-based techniques.

Other methods have also aimed to compute a surface by deforming a simple template while updating its connectivity \cite{wang2018pixel2mesh,pan2019deep}, fitting parameterized \cite{groueix2018papier,williams2019deep} or mesh-aware patches \cite{badki2020meshlet}, performing local (e.g., convex) shape decomposition. Majority of these schemes are restricted to particular shape topology or category and again do not necessarily guarantee point set preservation.

\begin{figure*}[t]
    \centering
    \includegraphics[width=\textwidth]{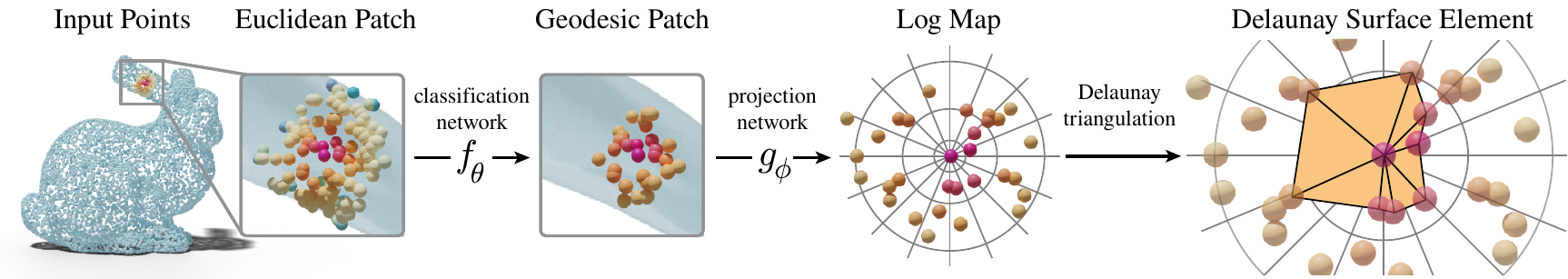}
    \caption{Overview of Delaunay Surface Element (DSE) generation. For any point $p_i$ in an input point cloud, we select the k-nearest neighbors and extract the subset of points that are in the geodesic neighborhood of $p_i$, using a learned classification network. A projection network then estimates a log map projection of the points into a 2D embedding, where we can apply Delaunay Triangulation to get a DSE.}
    \label{fig:overview}
\end{figure*}

\subsection{Learning mesh connectivity}
More directly, our work fits within the line of recent efforts aimed explicitly at learning the mesh connectivity for a given shape geometry. An early approach, Scan2Mesh~\cite{dai2019scan2mesh} developed a graph-based formulation to generate triangles in a mesh. However, the method uses a costly volumetric representation, does not aim to produce manifold meshes, and specializes on particular shape categories.

Most closely related to ours are two very recent approaches aimed directly to address the point set triangulation problem. The first method PointTriNet~\cite{sharp2020pointtrinet}  works on point clouds and, similarly to ours, uses a local patch-based network for predicting connectivity. However, this technique processes triangles independently and only promotes watertight and manifold structure through soft penalties. The second method was presented in \cite{liu2020meshing}, and estimates local connectivity by predicting the ratio between geodesic and Euclidean distances. This is a powerful signal, which is then fed into a non-learning based selection procedure, which aims to finally output a coherent mesh.

In contrast to both of these approaches \cite{sharp2020pointtrinet,liu2020meshing}, we formulate the meshing problem as learning of (local) Delaunay triangulations.
Starting from the restricted Voronoi diagram based formulation proposed in \cite{boltcheva2017surface} we use data-driven priors to directly learn local projections to create local Delaunay patches.
As a result, locally our network \emph{guarantees} the coherence of the computed mesh. As we demonstrate below, learning  Delaunay surface elements, both leads to better shaped triangles (i.e., more desirable angle distribution) and improves the overall manifold and watertight nature of the computed triangle mesh.

\section{Method}

We assume to be given an point set $P\in\mathbb{R}^{N\times3}$ sampled from a surface $\hat{S}$. Our goal is to create a mesh $M = (P', T)$  that approximates $\hat{S}$, by choosing a new triangulation $T$ that triangulates a subset $P'\subset P$ of the input point cloud.
It is easy to obtain a high-quality triangulation for any set of points that lies in \emph{2D}, via Delaunay triangulation~\cite{delaunay1934sphere}. However, when the set of points lies in 3D, finding a triangulation is a much harder problem.
A simple solution is to locally project points to an estimated tangent plane of the surface, resulting in local 2D embeddings where we can apply a Delaunay triangulation. However, this is problematic near complex geometry, such as edges or thin structures and is sensitive to an imperfect estimation of the tangent plane. \emph{Logarithmic maps}~\cite{do2016differential, herholz2019efficient}, or \emph{log maps} for short, provide a systematic solution to this problem by providing local geodesic charts of the ground truth surface that are good local parameterizations of complex geometry.

The core idea of our method is therefore to combine Delaunay triangulations and learned log maps to create small triangulated patches that we call \emph{Delaunay Surface Elements}~(DSEs). Each DSE approximates a small part of the surface and is guaranteed to have a manifold triangulation. Since neighboring log maps may disagree, especially in regions of high curvature, we align them locally with non-rigid transformations of the 2D parameterizations of each DSE.
%
DSEs enable us to maintain the good properties of Delaunay Triangulations, like manifoldness and high-quality triangles, within a data-driven approach, that learns to extract local geodesic patches and parameterize them with a the log map, thereby increasing robustness and reconstruction accuracy.
%
%
Our approach proceeds in four steps (the first two steps are illustrated in Figure~\ref{fig:overview}):
\begin{enumerate}
    \item For each point $p_i\in P$, a network estimates a geodesic ball, by extracting a 3D patch $P^i\in\mathbb{R}^{k\times3}$ made up of its $k$-geodesically-closest points.
    \item For each 3D point patch $P^i$, a second network approximates the log map parameterization, to get a 2D embedding of the patch, denoted $U^i\in\mathbb{R}^{k\times2}$.   
    \item We improve the consistency of neighboring patches by aligning their 2D embeddings, giving us improved patch embeddings $\hat{U}^i$, which we then use to compute the Delaunay Surface Elements.
    \item The Delaunay Surface Elements \emph{vote} for candidate triangles, which are then aggregated iteratively into a mesh.
\end{enumerate}

\subsection{Constructing Local Embeddings}
The first two steps in our method are aimed at creating a patch $P^i$ around each point $p_i$ and a local 2D embedding $U_i$ of the points inside the patch. These two ingredients will later be used to compute a 2D Delaunay triangulation that defines a Delaunay Surface Element.

\paragraph{Geodesic patch construction}
Given the point $p_i$ and its $K$ nearest neighbors $Q^i$, we train a network to find a subset of $k$ points from these neighbors that are \emph{geodesically} closest to $p_i$ on the ground truth surface. In our experiments, we set $K=120$ and $k=30$. More details on the choice of $k$ and $K$ are provided in Section \ref{sec:neighborhood_size} of the supplementary. The network is trained to model a function $c_j := f_\theta([q^i_j, d^i_j]\ |\ Q^i)$ that classifies each point $q^i_j$ in $Q^i$ as being one of the $k$ geodesically closest points if $c_j=1$ or not if $c_j=0$. We concatenate the Euclidean distance $d^i_j$ to the center point as additional input. The network is parameterized by $\theta$, conditioned on the point set $Q^i$, and models a function from 3D position to classification value. We train this network with an L2 loss $\|c_j - \sigma(\hat{c}_j)\|^2$, where $\hat{c}_j$ is the ground classification and $\sigma$ is the sigmoid function.
To obtain a fixed number of $k$ points, we select the top-$k$ points based on their predicted labels $c^i_j$, giving the (geodesic) patch $P^i$.

\paragraph{Log map estimation}
We train a second network to compute the log map coordinates of each point in $P^i$, denoted as $U^i \in \mathbb{R}^{k\times2}$.
The network is trained to model a function $u^i_j := g_\phi([p^i_j, d^i_j]\ |\ P^i)$, where $\phi$ denotes the network parameters and $p_j$ are the 3D coordinates of a point in $P^i$. These coordinates are concatenated with the Euclidean distance $d^i_j$ to the center point. The network outputs the log map coordinates $u^i_j$ in $U^i$, consisting of the Euclidean coordinates of the log map with an origin at the center point of the patch.
Like the classification network, this network is conditioned on the input point set $P^i$.
\changed{We use the sum of two losses: a loss that penalizes the difference to the ground truth coordinates and one that penalizes only the radial component of the coordinates, i.e. the geodesic distance to the center.} Since log map coordinates are defined only up to a 2D rotation around the central point, we use the Kabsh algorithm~\cite{berthold1987closed} to find  a 2D rotation and/or reflection that optimally aligns the predicted log map and the ground truth log map before computing the loss: $\|R U^i - \hat{U}^i\|_2^2$, where $\hat{U}^i$ is the ground truth and $R$ is the optimal rigid transformation computed by the Kabsh algorithm. Note that the Kabsh algorithm is differentiable. \changed{Our second loss measures the error in the radial component: $\sum_j (\|u^i_j\|_2 - \| \hat{u}^i_j\|_2)^2$. This loss measures how well the network can recover geodesic distances regardless of the orientation in the patch.}

\paragraph{Network architecture} When approximating log maps with a network, continuity is an important property. If the estimated mapping from 3D space to the 2D log map parameterization is not continuous, the resulting Delaunay triangulation may have flipped or intersecting triangles. We base our architecture on FoldingNet~\cite{yang2018foldingnet} that produces continuous mappings from an input to an output domain. Unlike the original implementation, however, which maps from 2D to 3D, we want to map from 3D to 2D.
%
Our experiments have shown that this network architecture leads to more continuous results than a PointNet-based architecture. We have also found that it improves the performance of our classification network, where we also adopt an architecture based on FoldingNet. Since we train our network on individual patches, we can train on relatively small datasets, where each shape provides a large number of patches as training samples.
More details on the architecture are provided in Section \ref{sec:architecture_details} of the supplementary.


\subsection{Combining Delaunay Surface Elements}
\begin{figure}[t]
    \centering
    \includegraphics[width=\columnwidth]{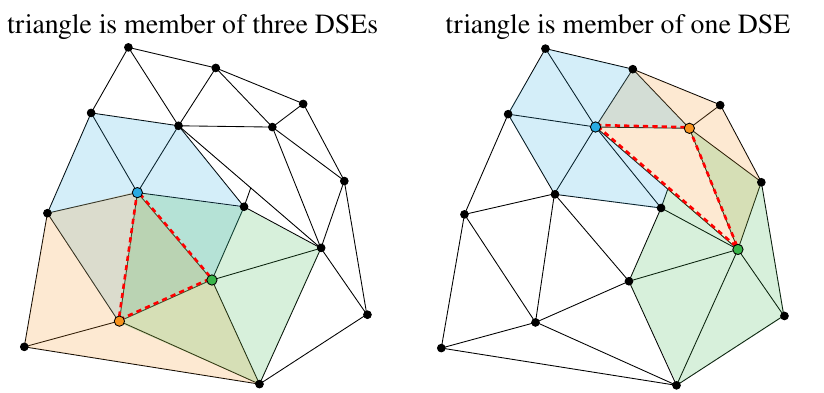}
    \caption{Triangle membership count. Delaunay Surface Elements are shown as colored triangles. Triangles that are part of exactly three DSEs, like the dotted red triangle on the left, result in a manifold triangulation. Triangles that are part of less than three DSEs, like the triangle on the right, result in non-manifold triangulations. We use this property to define a triangle confidence when selecting triangles.}
    \label{fig:selection}
\end{figure}

At this point, we have a local 2D parameterization for each patch. We could use these local parameterizations to construct a triangulation of the patch by Delaunay-triangulating it. However, each patch may be rather inconsistent with neighboring patches, in the sense that if two patches $P^i,P^j$ share three points $a,b,c$, the Delaunay triangulation of $U^i$ may produce the triangle $(a,b,c)$ while the triangulation of $U^j$ may not, since the points are laid out differently in each of the two parameterization. An example is shown in Figure~\ref{fig:selection}, right. Hence, the final pair of steps is aimed at improving the consistency between the different patch parameterizations of neighboring DSEs before combining all DSEs into the final mesh $M$.

\begin{figure}[t]
    \centering
    \includegraphics[width=\columnwidth]{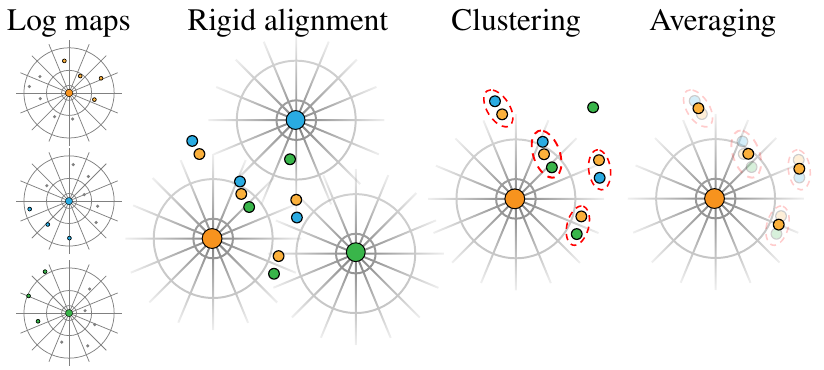}
    \caption{Log map alignment. To improve the consistency of log maps, we align corresponding points in neighboring log maps with rigid transformations. The resulting sets of corresponding points are then clustered to remove outliers and averaged, giving us 2D point embeddings that are more consistent with their neighbors.}
    \label{fig:alignment}
\end{figure}

\paragraph{Log map alignment} In 2D, Delaunay triangulation are guaranteed to produce a manifold triangulation. However, we produce independent 2D parametrizations for each DSE. Large differences in the parameterization of neighboring DSEs may make their triangulations incompatible (i.e., the union of their triangles may be non-manifold). In this step,
%
we locally align the log maps to one another to ensure better consistency, without requiring the construction of a global parameterization. Namely, a point $p_k\in P$ from the original point cloud has an image in the log maps of each patch that contains that point. We denote this set of all log map images of point $p_k$ as $R^k$. We say $U^i,U^j$ are neighbor patches if they both have a point in the same $R^k$. Denote the image of $p_k$ in the log map of each of the two patches as $U^i\left(p_k\right)$, $U^j\left(p_k\right)$, respectively.

Our approach is illustrated in Figure~\ref{fig:alignment}. Considering the patch $U^i$, we align the neighboring patch $U^j$ to it, by taking all corresponding points and using the Kabsch algorithm to find the rigid motion that best aligns (in the least-squares sense) the points based on their correspondences $U^i\left(p_k\right) \leftrightarrow U^j\left(p_k\right)$.
Repeating this for all  neighboring patches aligns them all to $U^i$. We then define the set $R^i_k$ to be the set of images of the point $p_k$ in the aligned log maps and cluster $R^i_k$ with DBSCAN~\cite{ester1996density}.
The largest cluster corresponds to the largest agreement between neighboring patches on the 2D coordinates $u^i_k$ of point $p_k$ in patch $i$. We  average  all 2D coordinates in the cluster to update $u^i_k$, and weigh the average based on the distance of each point in $R^i_k$ to the center of its patch.
Applying this process to all 2D coordinates $U^i$ in each patch, we
get a corrected log map $\hat{U}^i$ for each patch, giving us DSEs that are more consistent with the neighboring DSEs.

\paragraph{Delaunay triangulation}
Given a patch $P^i$ and its 2D parameterization $\hat{U}^i$, we can compute a Delaunay Triangulation on the 2D points $u^i_j$. If $\hat{U}^i$ approximates the log map, this gives us a manifold triangulation of the 3D patch that locally approximates the ground truth surface $\hat{S}$. We define a Delaunay Surface Element $D:=(P^i, T^i)$ as the set of Delaunay triangles $T^i$ corresponding to the Voronoi cell centered at $p_i$. These triangles form an umbrella with $p_i$ as its central point. We restrict our triangulation to triangles that include the central point, as triangulations are increasingly inconsistent with neighboring DSEs as the distance from the central point increases.

\paragraph{Triangle selection}
Combining the triangles of all DSEs yields a set of candidate triangles that we use in a final triangle selection step to obtain a near-manifold mesh. We base our selection criteria on our DSEs by observing that a triangulation is manifold exactly if all triangles are part of three DSEs (see Figure~\ref{fig:selection}).
%
Therefore, we divide our triangles into three confidence bins. Triangles that appear in three different DSEs will be considered the most likely to appear in a manifold triangulation. And triangles that appear only once are considered least likely. Finally, we use the triangle selection process proposed in \cite{liu2020meshing} to produce a triangulation based on our priority queue.

\section{Results}

We evaluate our method by comparing the quality and accuracy of our reconstructed meshes to the current state-of-the-art.

\paragraph{Dataset}
Since our networks are trained on individual patches, our method is able to train successfully from a small training set of shapes. Each shape provides a large set of patches as training samples. We create a dataset with a total of 91 shapes chosen from Thingi10k~\cite{Thingi10K} and the PCPNet~\cite{guerrero2018pcpnet} dataset, that we call \textsc{FamousThingi}, since the PCPNet dataset contains several shapes that are well-known in the graphics and vision literature. Each shape is sampled uniformly with 10k points. We compute ground truth log maps at each point using the recent method by Sharp et al.~\cite{Sharp:2019:VHM}.
The training set contains 56 of these shapes and the remaining shapes are used for evaluation. Example shapes and more details are given in
Section \ref{sec:dataset_examples} of
the supplementary.


\begin{table}
\begin{center}
\caption{\changed{Quantitative results on the \textsc{FamousThingi} testset. We compare the percentage of non-watertight edges (NW), the Chamfer distance (CD) and normal reconstruction error in degrees (NR).
}}
\begin{tabular}{|r|S[table-format=2.1]|S[table-format=2.1]|S[table-format=2.1]|}
\hline
Method & {NW (\%)} & {CD \footnotesize $*1^{e-2}$} & {NR} \\
\hline\hline
ball pivoting & 25.7 & 0.524 &  6.59
\\
PointTriNet \cite{sharp2020pointtrinet} & 17.2
 & 0.337 &6.24\\
RVE \cite{boltcheva2017surface} & 9.2 & 0.344 & 15.71\\
IER meshing \cite{liu2020meshing} & 5.3 & 0.343&6.30\\
$\alpha$-shapes $3\%$ & 2.5 & 0.939& 28.50\\
$\alpha$-shapes $5\%$ & 1.7 & 1.064& 17.69\\
Ours&\bfseries 0.4 & \bfseries 0.326&\bfseries 5.23\\


\hline
\end{tabular}
\end{center}
\label{tab:quant_comparison}
\end{table}

\begin{figure*}
\begin{center}
\includegraphics[width=1.0\linewidth]{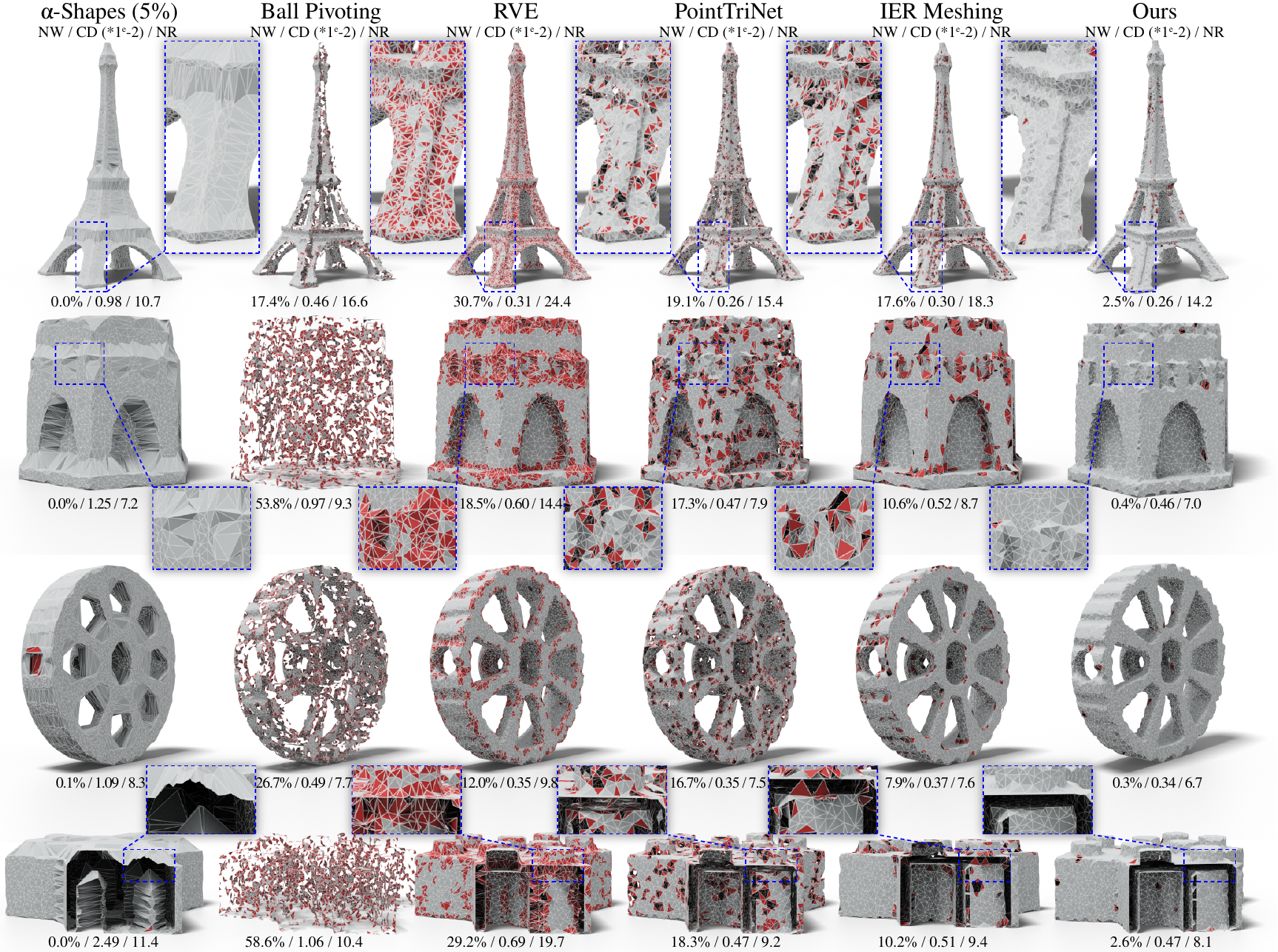}
\end{center}
   \caption{Qualitative comparison. We compare four meshes reconstructed by our method to the results of five current methods. Non-manifold triangles are marked in red and we show both the percentage of non-watertight edges (NW) and the Chamfer distance multiplied by 100 (CD) below each shape. Note that classical non-data-driven methods struggle to separate thin surfaces and data-driven methods have significantly more non-manifold triangles.
   }
\label{fig:qualitative}
\end{figure*}


\subsection{Comparison to Baselines}

We compare our method to recent state of the art learning based methods for point-set triangulation, as well as to more classical methods.

\paragraph{Ball pivoting~\cite{bernardini1999ball} and $\alpha$-shapes \cite{edelsbrunner1994three}.} These two classic techniques use the concept of rolling a ball on the surface to deduce connectivity at points of contact. For ball-pivoting, the ball radius is automatically guessed as the bounding box diagonal divided by the square root of the vertex count. For $\alpha$-shapes, we report two different choices of
the radius parameter $\alpha$, as 3\% and 5\% of the bounding box diagonal.

\paragraph{Restricted Voronoi estimation~\cite{boltcheva2017surface} (RVE)} This method is the closest existing baseline to our method. It estimates Voronoi cells restricted to the surface by projecting local patches to local tangent planes.  Note that this method requires normal information that we estimate from the input point cloud.

\paragraph{PointTriNet~\cite{sharp2020pointtrinet} and IER meshing~\cite{liu2020meshing}} We compare our method to two recent learning based methods for triangulating point clouds. We retrain PointTriNet on our dataset. Intrinsic-Extrinsic Ratio Guidance Meshing (IER meshing), however, needs a larger amount of data to train and overfits on our dataset. Since it is not patch based, it needs a larger variety of shapes to train. We use the pre-trained model provided by the authors, that was trained the larger ShapeNet dataset.

\paragraph{Metrics} We compare to these methods using two metrics for the mesh quality and two metrics for the mesh accuracy. \changed{As mesh quality measures, we use the percentage of non-watertight edges (NW) and the standard deviation ($A_\sigma$) of triangle angles in the mesh. Note that due to the triangle selection step, all the produced edges are manifold (have one or two adjacent triangles) but the edges can be open.} An angle of $60$ degrees corresponds to equilateral triangles, while skinny triangles have more extreme angles.

As a measure of the surface reconstruction accuracy, we use the Chamfer Distance~\cite{Barrow:1977:Chamfer, fan2017point} (CD) between a dense point set $P_M$ sampled on the reconstructed surface and a dense point set $P_{\hat{S}}$ sampled on the ground truth surface:
\begin{equation*}
\begin{split}
\text{CD}(P_M,  P_{\hat{S}})\ =\ & \frac{1}{N } \sum_{p_i \in P_M} \min_{q_j \in P_{\hat{S}}} \|p_i-q_j\|_2\ + \\
& \frac{1}{N} \sum_{q_j \in P_{\hat{S}}} \min_{p_i \in P_M} \|q_j-p_i\|_2
\end{split}
\end{equation*}
\changed{
We also compare the normal reconstruction error (NR). At each vertex of the mesh we measure the angle difference in degrees between the ground truth normal and the normal obtained from our reconstructed mesh. }
\paragraph{Quantitative Comparison} In Table~\ref{tab:quant_comparison}, we show a quantitative comparison between our method and the baselines.
Our method yields lower chamfer distance, and less non-manifold edges, showing we both better-approximate the surface while at the same  time outputting a triangulation with far less non-manifold artifacts. Indeed, only the classic technique of $\alpha$-shapes manages to come close to our degree of manifoldness, at the cost of lower accuracy, due to filling in concave surface regions (see examples in Figure~\ref{fig:qualitative}).

\begin{figure}[t!]
\begin{center}
   \includegraphics[width=1.0\linewidth]{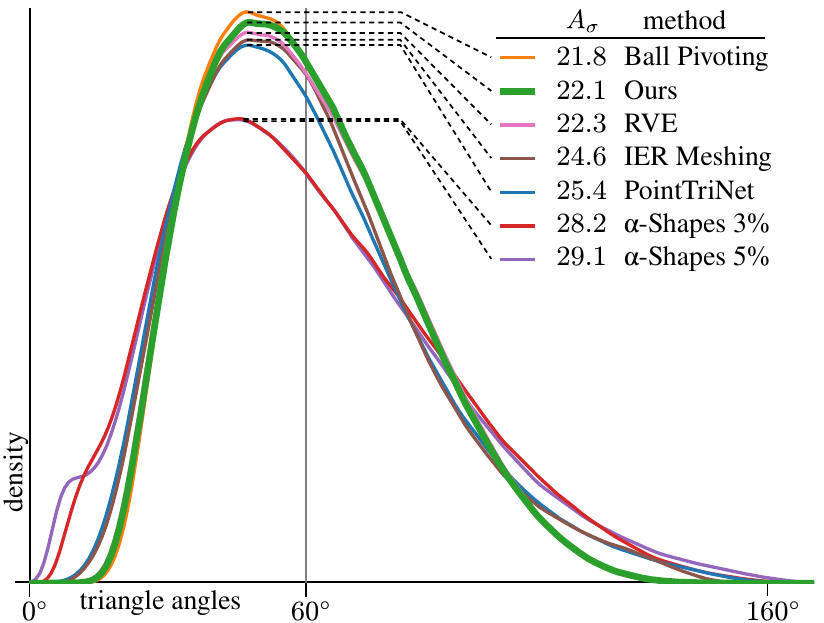}
\end{center}
   \caption{Distribution of triangle angles in the reconstructed meshes. Our method produces better shaped triangles than all other methods except for ball pivoting which sacrifices mesh manifoldness. We show the angle variance next to each method.}
\label{fig:triangle_angles}
\end{figure}

In Figure~\ref{fig:triangle_angles}, we evaluate the quality of the generated triangles by considering the histogram of triangle angles. The standard deviation of each method is given next to its name. Our method yields superior triangle quality to all learning-based methods, and to all classic techniques except for ball-pivoting, which achieves better triangle quality by sacrificing manifoldness to a large degree.

\begin{figure}
\begin{center}
\includegraphics[width=1.0\columnwidth]{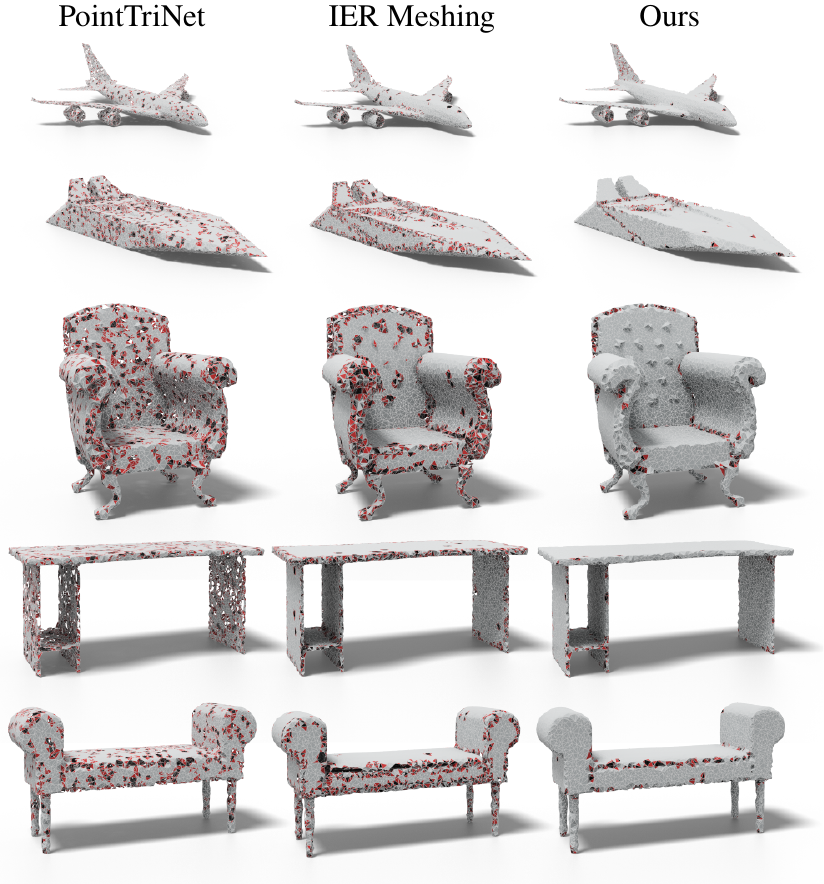}
\end{center}
   \caption{Qualitative comparison on ShapeNet~\cite{shapenet2015}. We compare with the two data-driven methods PointTriNet and IER Meshing on five shapes taken from five different categories of the ShapeNet dataset. Our approach results in more manifold meshes, especially in detailed areas like the backrest of the chair.
   }
\label{fig:shapenet}
\end{figure}

\paragraph{Qualitative Comparison}  We show qualitative results in Figure \ref{fig:qualitative}, on 4 meshes of our \textsc{FamousThingi} dataset. Non-manifold triangles are visualized in red, with the percentage of non-manifold triangles, as well as the Chamfer distance error, written beneath each result. The figure gives a very clear visual insight to the numbers from Table \ref{tab:quant_comparison}: the classic techniques work in a non-adaptive way which enables them to produce meshes with mostly-manifold edges, but they cannot handle thin and tight structures, like the scaffolds of the tower. In contrast, the learning-based methods are more local and can handle the concavities in, e.g., the wheel, but fall short on producing manifold triangulations. Our method, combining the robustness of classic Delaunay triangulation, with modern, data-driven learning techniques, manages to produce triangulations that both respect the original fine geometry and have less non-manifoldness.

We show additional results on five shapes of the ShapeNet dataset~\cite{shapenet2015} in Figure~\ref{fig:shapenet}. Compared to the two data-driven methods PointTriNet and IER Meshing, we improve upon the manifoldness, especially in regions with detailed geometry and high curvature, like the edge of the table, or the backrest of the chair. The results show a similar trend as in our \textsc{FamousThingi} dataset. Note that IER meshing is trained on the ShapeNet dataset while PointTriNet and our method are trained on \textsc{FamousThingi} dataset, demonstrating the ability of our method to generalize to unseen data. We show quantitative and qualitative results on the ShapeNet dataset in
Section \ref{sec:shapenet} and additional qualitative results in \ref{sec:more_qualitative_results} of
the supplementary.

\paragraph{Limitations} \changed{ Finding a geometrically complex surface, like on parts of the Eiffel tower in Figure \ref{fig:qualitative}, can be difficult. In such cases, the geodesic neighbors or logmap networks may misclassify/misplace some points. Moreover, thin parts of a model are particularly challenging. We can handle these cases better than existing works (Lego piece of Figure \ref{fig:qualitative}, or the plane wing in Figure \ref{fig:shapenet}. More extreme cases, like the leafs of a plant, would require training on a dataset where these cases are more common. }
\paragraph{Non uniform sampling} We evaluate our method on non uniformly sampled point clouds. In particular we sample points following a probability gradient along the y-axis (horizontal). We observe in Figure \ref{fig:teaser} (bottom) that our method performs better than other learning-based baselines. Note that PointTriNet, IER Meshing and our method have \emph{not} been retrained on a non-uniformly sampled dataset. We provide further evaluation on non uniformly sampled point clouds in
Section \ref{sec:non_uniform} of
the supplementary.

\subsection{Ablation study}

\begin{table}
\begin{center}
\caption{\changed{Ablation study over the components of our method. Log map alignment, triangle selection as well as the
the log map
parametrization improve manifoldness in the output meshes.}}
\begin{tabular}{|r|S[table-format=2.1]|S[table-format=2.1]|S[table-format=2.1]|}
\hline
Method & {NW (\%)} & {CD \footnotesize $*1^{e-2}$} & NR \\
\hline\hline

Ours w/o align, select & 22.51 &0.326&7.26\\
Ours w/o select & 10.98 & 0.348&6.86\\
Ours w/o log maps & 1.18 & 0.334&5.93\\
Ours w/o align & 1.07 & \bfseries 0.325&\bfseries 5.19\\
Ours & \bfseries 0.40 &  0.326&5.22\\

\hline
\end{tabular}
\label{tab:ablation}
\end{center}
\end{table}

\changed{
We evaluate the impact of each step in our pipeline using an ablation study, shown in Table \ref{tab:ablation}.
We remove one component at a time and compute the percentage of non-watertight edges (NW), Chamfer distance (CD) and normal reconstruction error (NR) as described before.
We first remove the \emph{align}ment of the logmaps of the delaunay triangulation, which results in a slight degradation manifoldness. Next, we evaluate the efficacy of our triangle \emph{select}ion process by instead creating a mesh from all triangles in our Delaunay Surface Elements. This results in a significant drop in manifoldness of the triangulation, since we do not achieve perfect alignment of our logmaps. Dropping both the alignment and the selection results in a much more significant decrease in manifoldness than just removing the selection -- this hints that the alignment is indeed producing more consistent local DSE's. Lastly, we replace the \emph{log map}s with simple 2D projections, to get the local patch parameterization along the approximated normal vector. Please note that we still use the learned geodesic neighborhood. Manifoldness deteriorates as well, showing the necessity of our specific parameterization method. In particular, the 2D projection parametrization performs poorly for complex shapes such as the Eiffel tower (NW: $5.59\%$ (w/o logmaps), $2.48\%$ (Ours)) or Trilego (NW: $5.59\%$ (w/o logmaps) NW: $1.64\%$ (Ours)) shapes.
Note that the Chamfer distance is not significantly increased by the removal of any component from our pipeline, as our method's locality prevents strong errors in the surface location by design, due to considering only the learned geodesic neighborhoods of the surface. We provide additional ablation of the learned Logmap component in Section~\ref{sec:logmap_ablation}  of the supplementary.}





\section{Conclusion}

We presented Delaunay Surface Elements for robust surface reconstruction from points sets.  In the process, we combine the best of two worlds: 2D Delaunay triangulation from classical computational geometry which comes with guarantee about mesh quality and manifoldness; and local logmaps learned using networks, followed by synchonization to get local data-driven 2D projection domains to handle non-planar regions. We demonstrated that the method can be trained with very limited training data and produces near-manifold triangulations that respect the original point set and have a higher mesh quality than the state-of-the-art.

In the proposed method, the final mesh extraction is done via a non-differentiable growing approach. In the future, it would be interesting to also learn the triangle selection via a network. This could enable a truly end-to-end optimization and allow us to optimize for context-specific point distributions accounting for data-priors (e.g., sharp edges) and scanner characteristics. Another direction would be to consider weighted Delaunay triangulations that provide additional freedom to local triangulations. 

\section{Acknowledgements}

Parts of this work were supported by an Adobe internship, the KAUST OSR Award No. CRG-2017-3426, the ERC Starting Grant No. 758800 (EXPROTEA) and the ANR AI Chair AIGRETTE.

{\small
\bibliographystyle{ieee_fullname}
\bibliography{main_learningDSurfElements.bbl}
}



\appendix

\section{Non-uniform sampling}
\label{sec:non_uniform}

\begin{figure}
\begin{center}
\includegraphics[width=1.0\columnwidth]{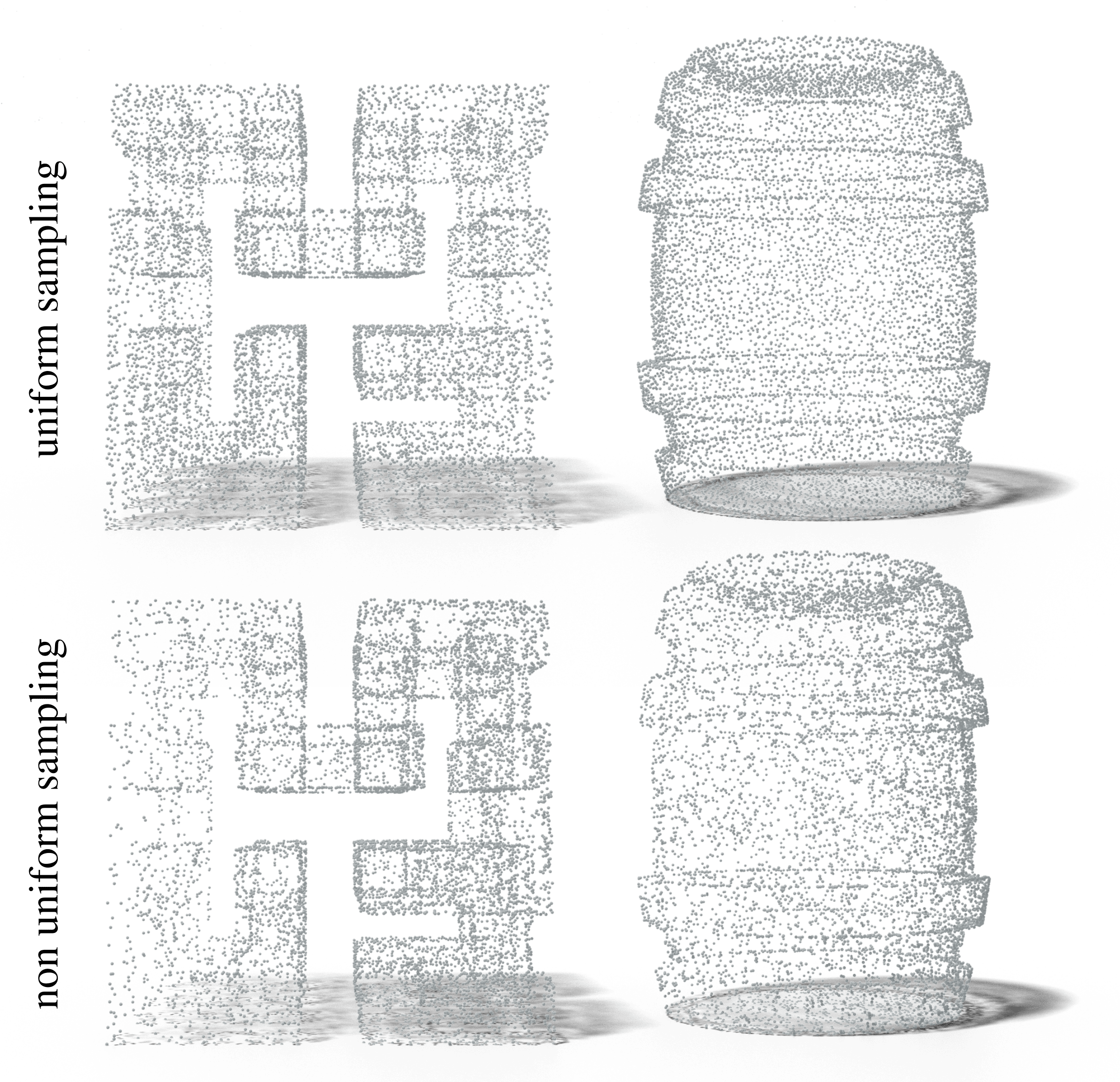}
\end{center}
   \caption{Examples of uniformly sampled point clouds (top) and non uniformly sampled point clouds. The density of points follows a gradient along the y axis (horizontal). }
\label{fig:non_uniform_pc}
\vspace{-10pt}
\end{figure}

\begin{figure}
\begin{center}
\includegraphics[width=1.0\columnwidth]{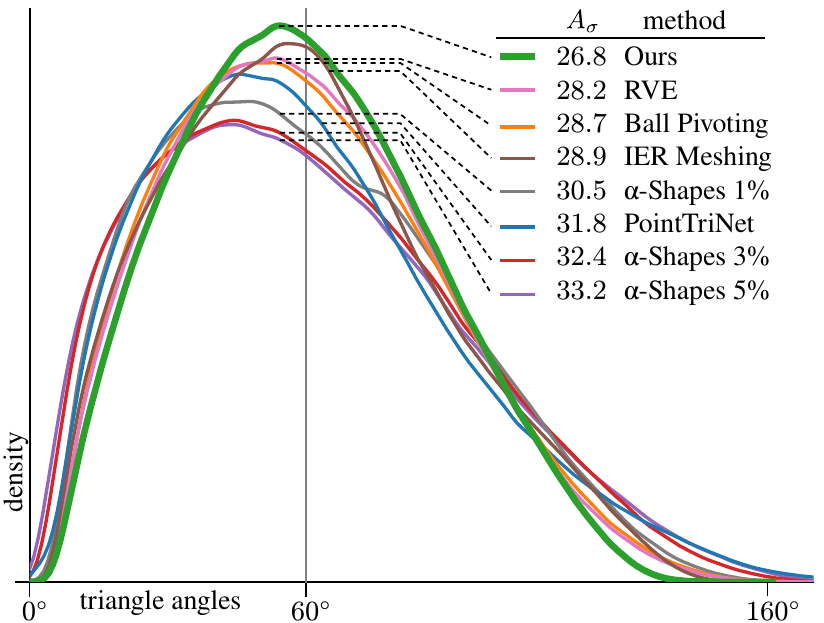}
\end{center}
   \caption{Triangle angles distribution. Our method produces triangles with angles more centered around 60 degrees and fewer very obtuse or very acute angles. }
\label{fig:angles_distrib}
\vspace{-10pt}
\end{figure}

We provide additional results on a non-uniformly sampled variant of the \textsc{FamousThingi} dataset.
We sample points following a density gradient along the y-axis (horizontal in the figures), where point density correlates with the y-coordinate. A few examples are shown in Figure~\ref{fig:non_uniform_pc} (bottom).
We did \emph{not} retrain on this dataset variant and evaluate the same model we used for the uniform point clouds.
In Table~\ref{tab:non_uniform} we show that our method remains robust even with this non-uniform sampling, with only a small decrease in performance compared to uniform sampling. IER meshing takes the largest performance hit with over twice as many non-manifold triangles and significantly increased Chamfer distance. Overall our method shows a similar improvement over the baselines as in uniform sampling.
The angle distribution of triangles produced by our method is compared to all baselines in Figure~\ref{fig:angles_distrib}. Our method achieves the best performance with angles more centered around 60 degrees.

\begin{figure*}[t]
\begin{center}
\includegraphics[width=1.0\linewidth]{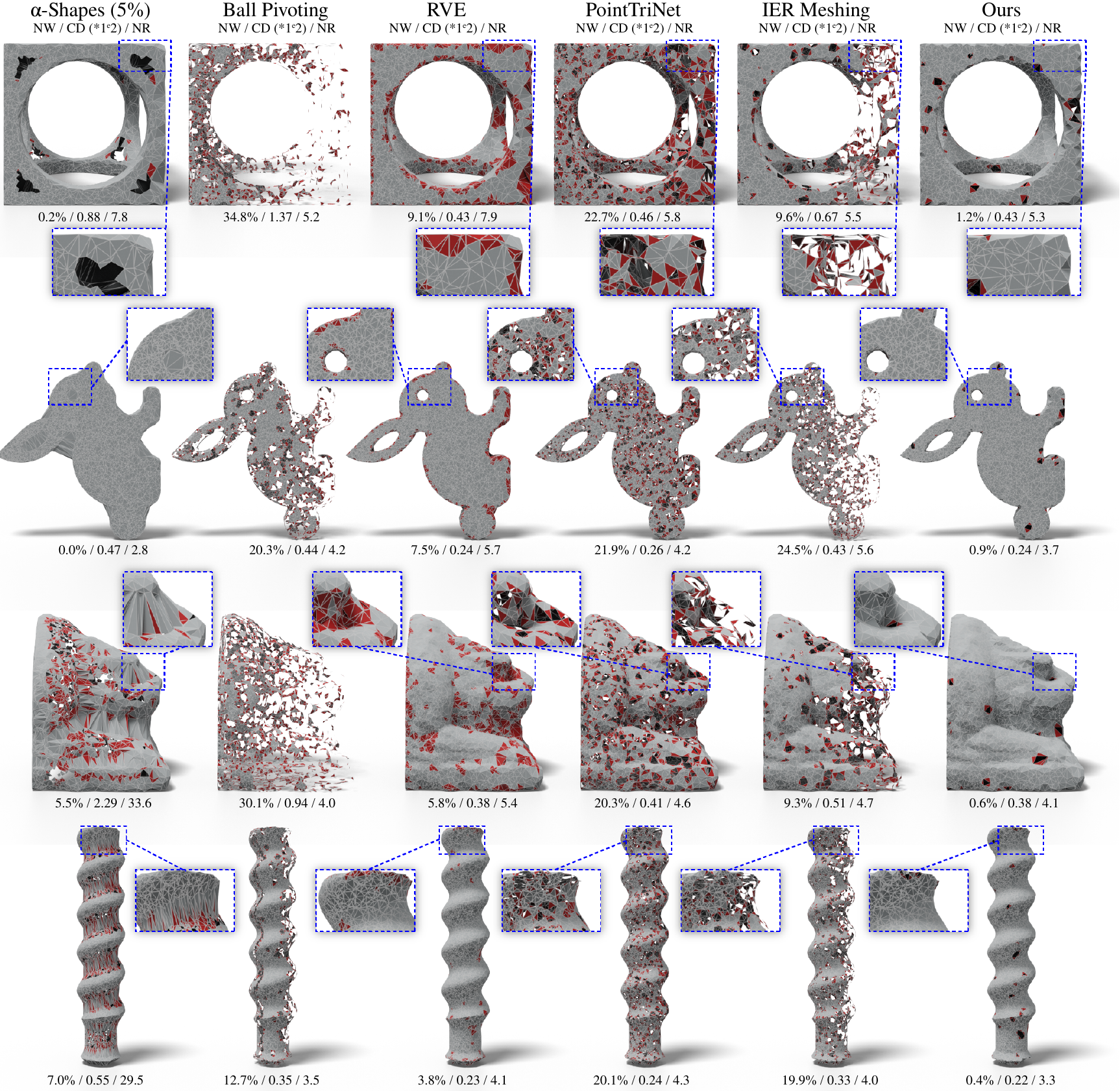}
\end{center}
   \caption{Surface reconstructions from non-uniform point clouds. Non-manifold triangles are marked in red. Shapes are sampled more densely to the left and more coarsely to the right. We can see that methods struggle to reconstruct the coarsely sampled parts of the point cloud. While our method also has slightly more errors in the coarsely sampled regions, the mesh quality drops by a much smaller amount from densely to coarsely sampled regions.}
\label{fig:non_uniform_reconstruction}
\vspace{70pt}
\end{figure*}

We show qualitative comparison in Figure \ref{fig:non_uniform_reconstruction}. We observe that ball pivoting and IER meshing are particularly impacted by the non uniform sampling while our method achieves the best quality reconstructions.

\begin{table}[h]
\begin{center}
\caption{\changed{Quantitative comparison the \textsc{FamousThingi} testset where points are sampled non-uniformly. We compare the percentage of non-watertight edges (NW), the Chamfer distance (CD), and the normal reconstruction in degrees (NR) to all baselines.}}
\begin{tabular}{|r|S[table-format=2.1]|S[table-format=2.1]|S[table-format=2.1]|}
\hline
Method & {NW (\%)} & {CD \footnotesize $*1^{e-2}$} & NR\\
\hline\hline
Ball pivoting & 31.5 & 0.396 &6.84\\
PointTriNet  & 14.2  &  0.383 & 6.59\\
IER meshing  & 13.5&  0.487 & 7.00\\
RVE & 11.0 & 0.396 & 9.08\\
$\alpha$-shapes 1\%  &3.5  & 3.228 &  63.21\\
$\alpha$-shapes 3\%  &2.7  & 0.971  & 28.88\\
$\alpha$-shapes 5\%  &1.7  & 1.061&  17.71\\
Ours  &  \bfseries 1.3 & \bfseries 0.356 &\bfseries 6.02\\

\hline
\end{tabular}
\label{tab:non_uniform}
\end{center}
\vspace{-10pt}
\end{table}

\section{Results on ShapeNet}
\label{sec:shapenet}

We compare our method to PointTriNet and IER meshing. Both our method and PointTriNet are trained on the \textsc{FamousThingi} dataset, showing their generalization performance, while IER meshing was trained on ShapeNet (since IER meshing requires more shapes for training than the other two methods). Even though this gives IER meshing an advantage, we observe in Table~\ref{tab:shapenet} that our method still produces shapes with better manifoldness and Chamfer distance than other methods.

\begin{figure*}[t]
\begin{center}
   \includegraphics[width=1.0\linewidth]{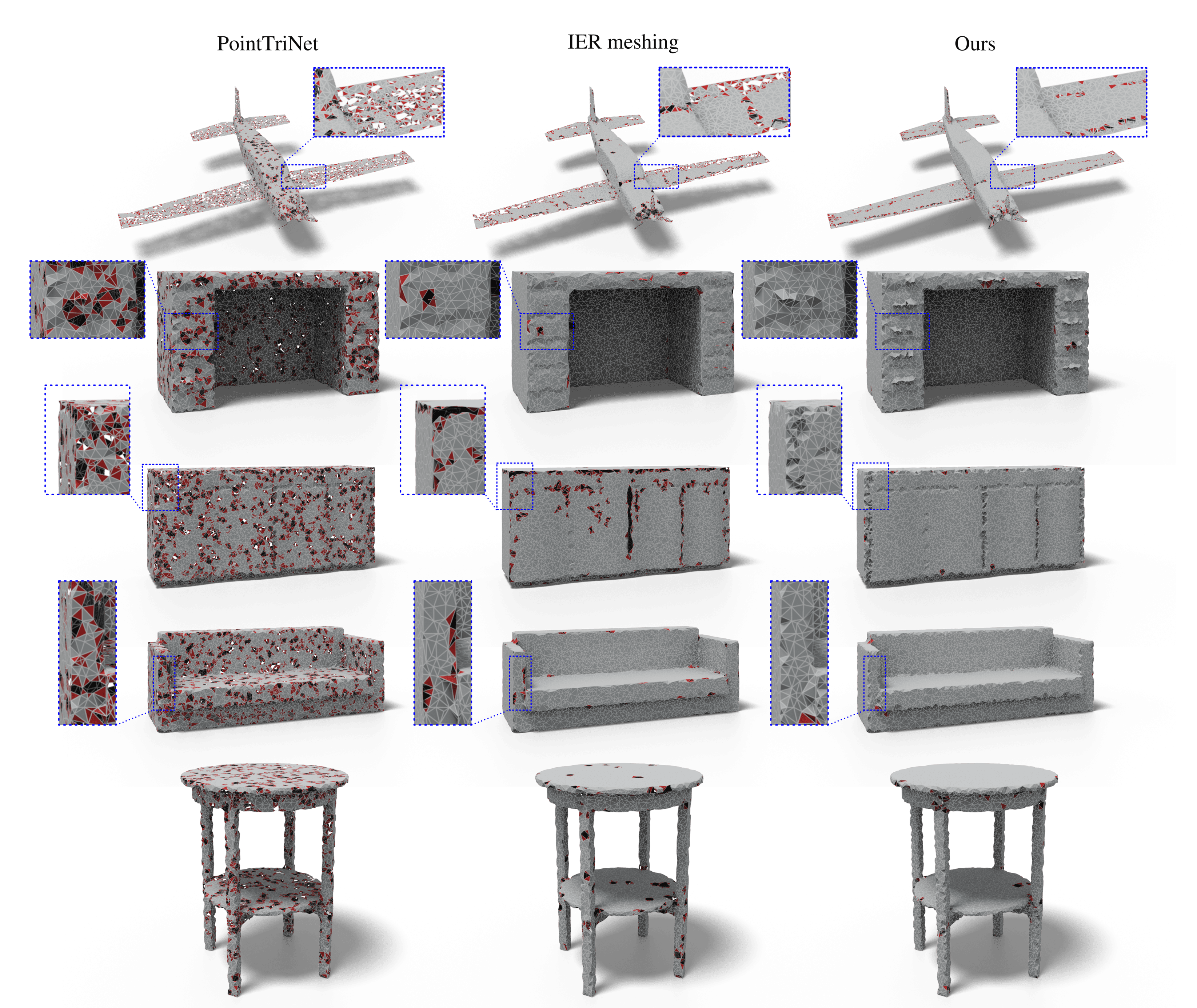}
\end{center}
   \caption{Qualitative results on ShapeNet testset. We do not retrain our method on the ShapeNet dataset while IER meshing \emph{was} trained on this dataset. Even so, our method produces more manifold meshes and preserves details such as the drawer handles (row 2) more accurately. We better separate the two sides (top and bottom) of the plane wings (row 1). Finally, our method presents fewer large holes in the reconstructed mesh.}
\label{fig:shapenet_qual}
\vspace{150pt}
\end{figure*}

Additional qualitative results are provided in Figure~\ref{fig:shapenet_qual}. We observe that our method produces meshes with better manifoldness and preserves details such as the drawer handles (row 2) or the two sides of the plane wings (row 1) more accurately. Finally, our method produces fewer large holes in the reconstructed mesh.

\begin{table}
\begin{center}
\caption{\changed{Quantitative comparison on 100 random shapes from ShapeNet. 
We compare our three main metrics to learning-based baselines. IER meshing was specifically trained on ShapeNet, while our method trained on a different dataset (\textsc{FamousThingi}). Even with this handicap, our method obtains better manifoldness and Chamfer distance.}}
\begin{tabular}{|r|S[table-format=2.1]|S[table-format=2.1]|S[table-format=2.1]|}
\hline
Method & {NW (\%)} & {CD \footnotesize $*1^{e-2}$}& NR \\
\hline\hline
PointTriNet  &22.33& 0.416& 10.95\\
IER meshing  &  6.96&  0.456& \bfseries 6.54\\
Ours  & \bfseries 5.51 & \bfseries 0.396 & 9.44\\
\hline
\end{tabular}
\label{tab:shapenet}
\end{center}
\vspace{-20pt}
\end{table}

\begin{figure*}[t!]
\begin{center}
   \includegraphics[width=1.0\linewidth]{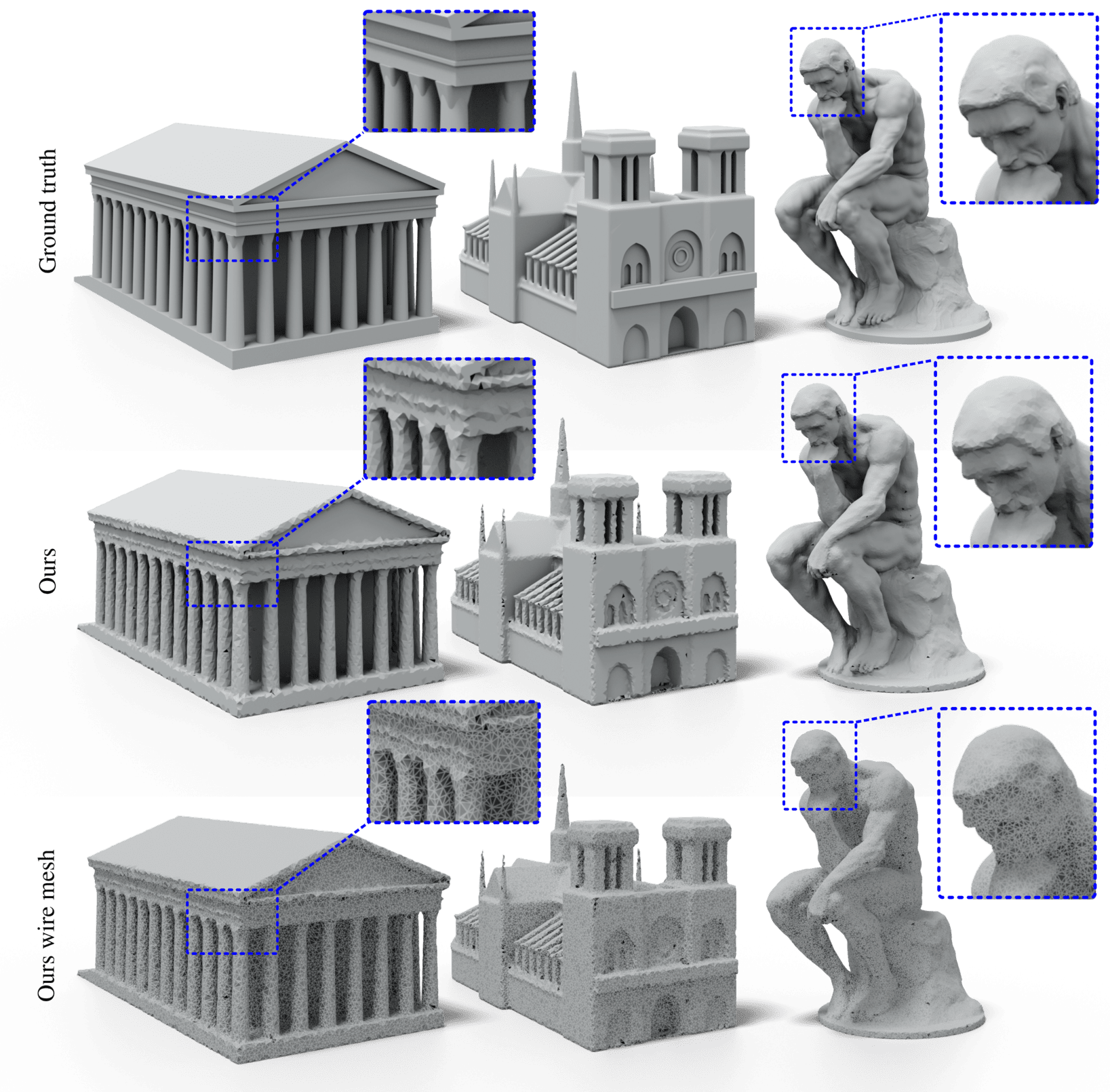}
\end{center}
   \caption{Meshing well-known monuments. We show the ground truth (top), the reconstructed mesh (middle), the reconstructed mesh with non manifold triangles colored in red (bottom). Our method generalizes well this more complex data that is also sampled more densely than our training set.}
\label{fig:monuments}
\vspace{80pt}
\end{figure*}

\begin{figure}[t]
\begin{center}
   \includegraphics[width=1.0\linewidth]{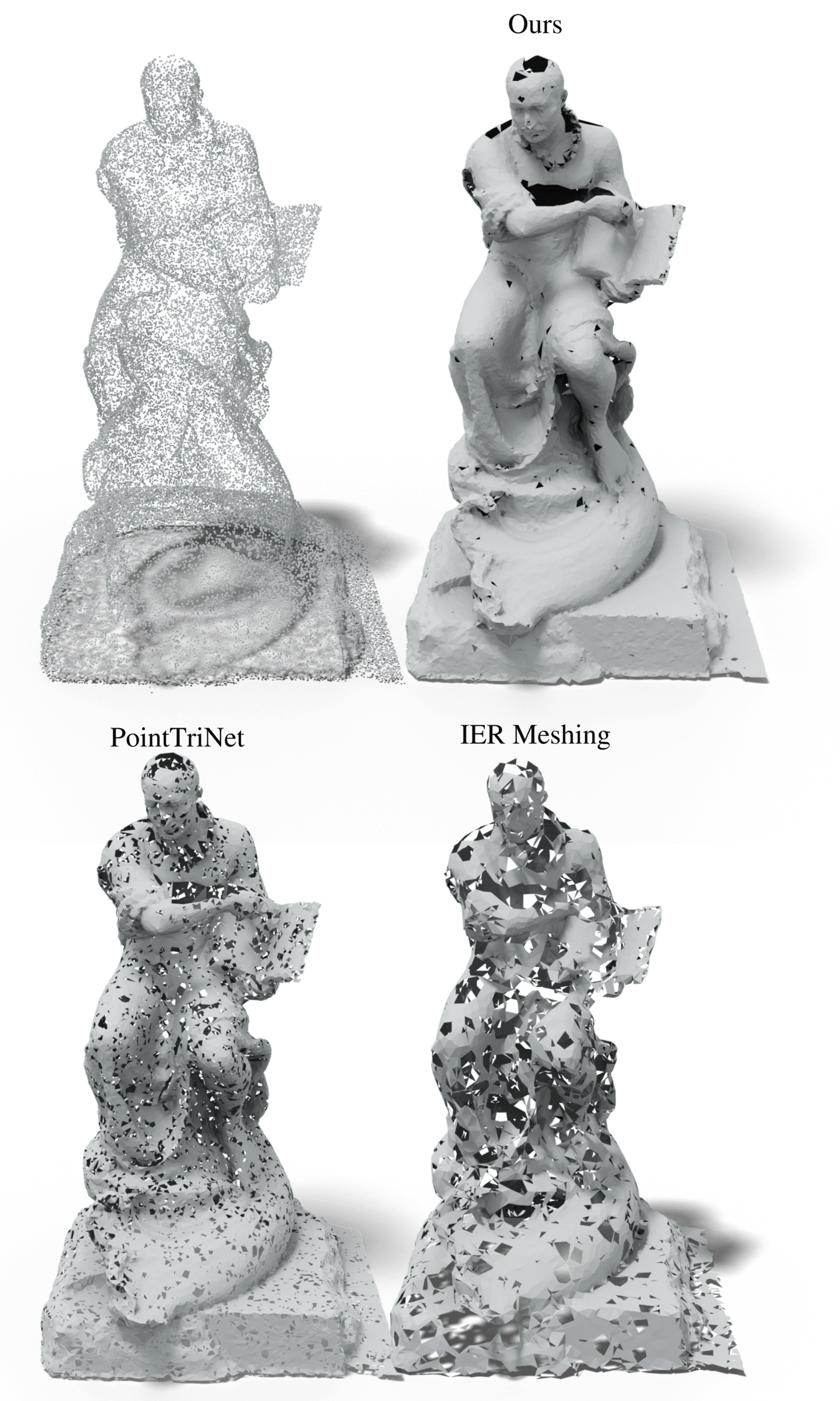}
\end{center}
   \caption{Reconstructing real scans from Tanks and Temples~\cite{Knapitsch2017}}
\label{fig:tanks_and_temples}
\end{figure}

\section{More Qualitative Results}
\label{sec:more_qualitative_results}
We provide additional qualitative results by meshing point clouds of well-known monuments obtained from \href{https://cults3d.com/en/3d-model/architecture/famous-paris-buildings}{Famous Paris Buildings}. Results are shown in Figure~\ref{fig:monuments}. Since these shapes are geometrically more complex than the shapes in \textsc{FamousThingi} or ShapeNet, we uniformly sample 50k points from each monument. We do not re-train on this dataset. Our method generalizes well to unseen data and denser point clouds.

\changed{
We also include a real scan reconstruction in Figure~\ref{fig:tanks_and_temples}. We reconstruct a point cloud with 50k sampled points and compare to other learning-based methods.
Since IER meshing can not handle 50k points,
we sample 12k points for comaring to IER meshing.}

\begin{figure}[t!]
\begin{center}
\includegraphics[width=1.0\columnwidth]{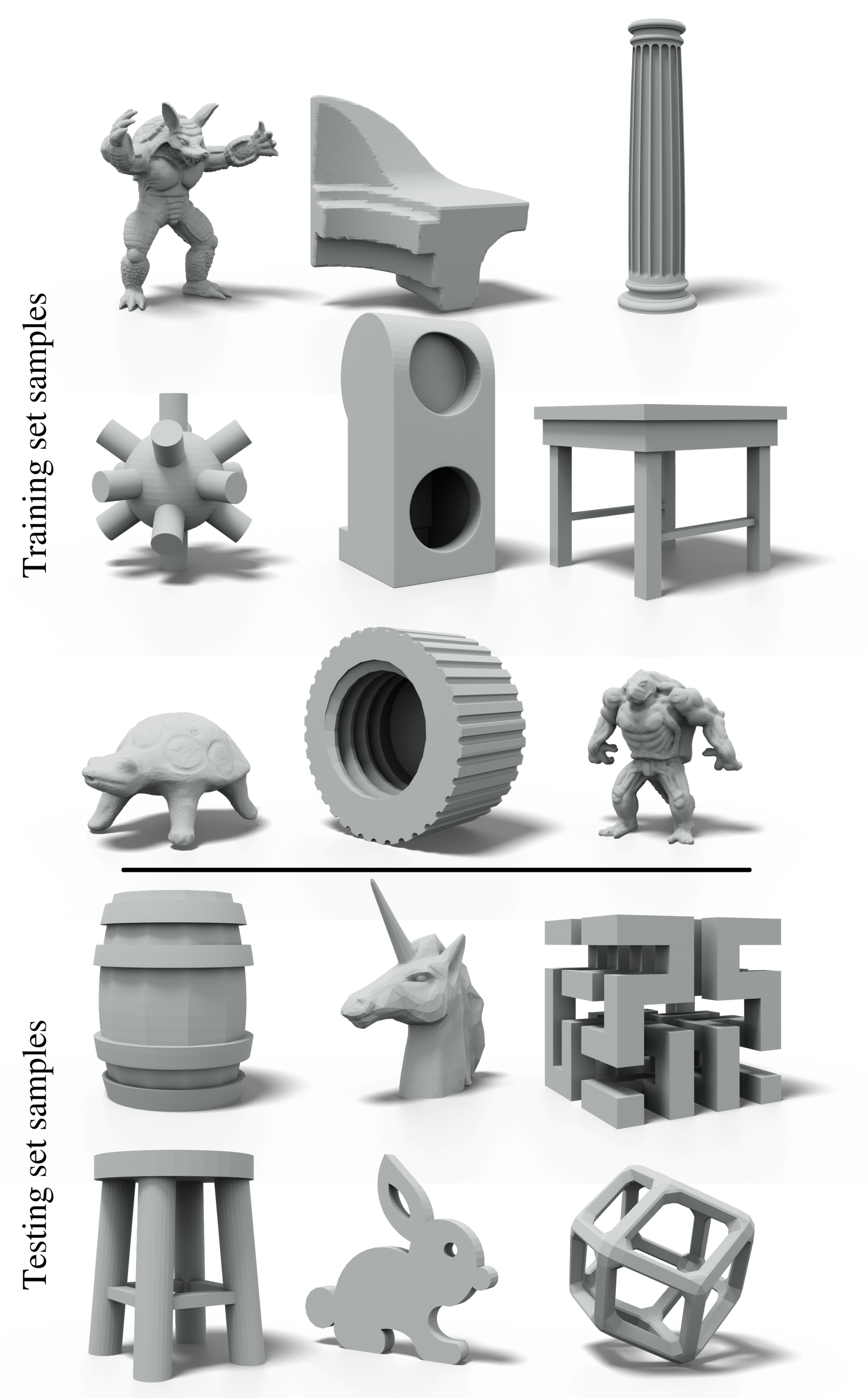}
\end{center}
   \caption{Examples from our dataset. We show ground truth meshes from both the training set and the test set.}
\label{fig:dataset}
\end{figure}

\section{Dataset Examples}
\label{sec:dataset_examples}

A few examples of shapes from our \textsc{FamousThingi} dataset are shown Figure~\ref{fig:dataset}. In Figure~\ref{fig:non_uniform_pc}, we show two examples of uniformly sampled point clouds we use as input to our method, and two non-uniformly sampled point clouds that we use in the experiments described in Section~\ref{sec:non_uniform}.

\section{Architecture Details}
\label{sec:architecture_details}

\begin{figure*}[t]
\begin{center}
\includegraphics[width=1.0\textwidth]{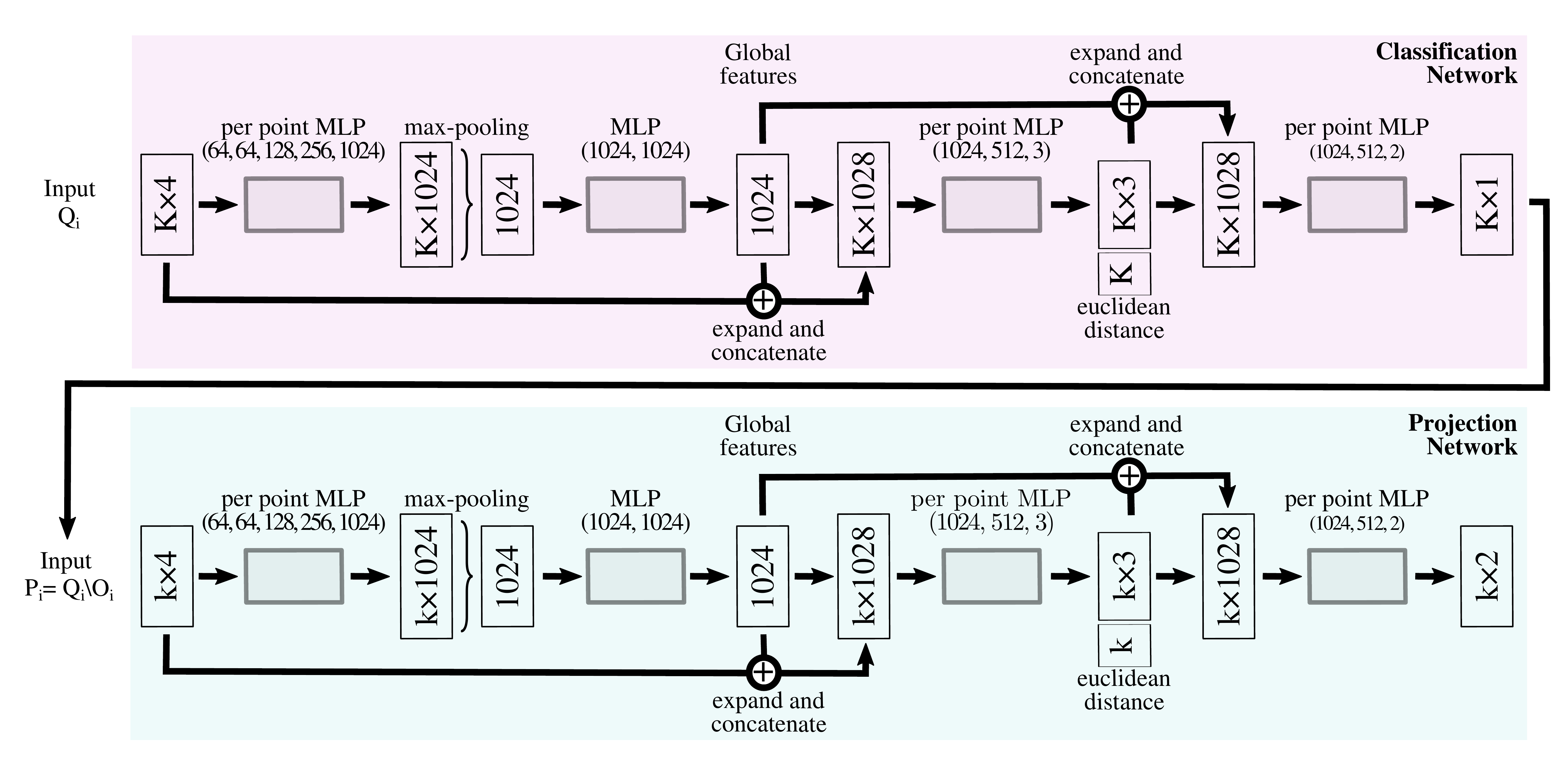}
\end{center}
   \caption{Detailed architecture of our pipeline. We first select a small geodesic patch using the classification network (purple). The projection network (blue) then applies a 2D projection to this patch that approximates a log map.
   }
\label{fig:architecture}
\end{figure*}

We show the detailed architecture of our geodesic patch \emph{classification network} and the 2D log map \emph{projection network} in Figure~\ref{fig:architecture}.
The classification network implements a function $c_j := f_\theta([q^i_j, d^i_j]\ |\ Q^i)$ that classifies if each point $q_i$ in the euclidean patch $Q^i$ is part of the geodesic patch $P_i$, while the projection network implements a function $u^i_j := g_\phi([p^i_j, d^i_j]\ |\ P^i)$ that projects points $p^i_j$ in the geodesic patch $P_i$ to their 2D log map coordinates $u^i_j$. Here $d_i$ is the euclidean distance from a point to its patch center.

Both networks use the same architecture based on FoldingNet~\cite{yang2018foldingnet}, except for their output dimension. They take as input a 3D point concatenated with the distance to the patch center and proceed to compute a $1024$-dimensional global feature vector for the input patch with a PointNet~\cite{qi2017pointnet}. Each input point is then augmented with this global feature vector and transformed by two blocks of per-point MLPs into a one-dimensional (classification network) or two-dimensional (projection network) per-point output.


\clearpage

\section{Runtime}
\label{sec:runtime}

\changed{We measure the average runtime of our method on point clouds of 10k vertices in Table~\ref{tab:runtime}, including the runtime for each step of the method.}

\begin{table}
\begin{center}
\caption{Average runtime estimation per step on 10k point clouds in seconds.}
\begin{tabular}{|l|l|l|l|}
\hline
 Log map est.& Log map align. & Selection & Total \\
\hline
 5.8& 24.8 &2.1  &32.8\\
\hline
\end{tabular}
\label{tab:runtime}
\end{center}
\end{table}


\section{Ablation of the Learned Logmap}
\label{sec:logmap_ablation}

\changed{We compare the performance of our learned logmap component to two baselines. The first baseline approximates the logmap by projecting neighboring points onto the tangent plane computed from the ground truth normal. The second baseline is the approach proposed in~\cite{gopi2000surface}, where points are rotated onto the tangent plane. Please note the both of these baselines use ground truth normal information, while our method does not. We evaluate the methods on a subset of 33 manifold shapes from our \textsc{FamousThingi} testset and sample 2k patches of k=30 geodesic neighbors per shape. We measure the MSE of the geodesic distance and of the 2D coordinates after patch alignment. Our method produces significantly better logmap estimates compared to other baselines as we show in Table~\ref{tab:learned_logmap}. }

\begin{table}
\begin{center}
\caption{Quantitative comparison our our learned logmap component to two logmap approximation methods.}
\begin{tabular}{|l|l|l|}
\hline
 & geodesic distance \footnotesize $*1^{e-3}$& 2D position  \footnotesize $*1^{e-2}$\\
\hline
Projection &1.943&2.627 \\
Rotation &1.943&2.835\\
Ours &\textbf{0.471}&\textbf{0.835}\\
\hline
\end{tabular}
\label{tab:learned_logmap}
\end{center}
\end{table}


\section{Ablation of Neighbor Counts $k$ and $K$}
\label{sec:neighborhood_size}

\changed{We evaluate our method on different values of the geodesic neighbor count $k$ ($20$, $30$, and $50$) and different values for the euclidean neighbor count $K$ ($80$, $120$, $160$) in Table~\ref{tab:ablation_k}. For each pair of $(k, K)$ values, we train our models for 30 epochs. The choice of the geodesic neighbor count $k$ affects the performance of our method significantly. If $k$ is small, the Delaunay element approximation quality is affected. If $k$ is large, it is more difficult for the logmap estimation network to produce a usable logmap. Changes in the choice of the euclidean neighbor count $K$ lead to less significant performance drops. In our experiment we choose the parameter values $k=30$ and $K=120$ which produce the best results for the non-watertightness and normal reconstruction metrics. Please note that for $k=30$ the difference in Chamfer distance values is negligible.}

\begin{table}
\begin{center}
\caption{Ablation of different values for the geodesic and euclidean and neighbor counts $k$ and $K$.}
\begin{tabular}{|c|c|c|c|c|}
\hline
k&K&CD(\footnotesize $*1^{e-2}$)&NW(\%)&NR\\ \hline
 20& 80&0.3437&5.569&5.921\\
20& 120&0.3394&4.381&5.845\\
20& 160&0.3496&4.712&6.483\\\hline
30& 80&0.3274&0.509&5.682\\
30& 120&0.3276&\textbf{0.485}&\textbf{5.661}\\
30& 160&\textbf{0.3272}&0.524&5.690\\\hline
50& 80&0.3335&1.822&6.046\\
50& 120&0.3282&0.667&5.856\\
50& 160&0.3286&0.728&5.883\\
\hline
\end{tabular}
\label{tab:ablation_k}
\end{center}
\end{table}



\end{document}